# Chatbot: A Conversational Agent employed with Named Entity Recognition Model using Artificial Neural Network

Nazakat Ali

Department of Electronics, Quaid-i-Azam University, Islamabad, Pakistan

nazakat5@gmail.com

**Abstract**

*Chatbot is a technology that is used to mimic human behavior using natural language. There are different types of Chatbot that can be used as conversational agent in various business domains in order to increase the customer service and satisfaction. For any business domain, it requires a knowledge base to be built for that domain and design an information retrieval based system that can respond the user with a piece of documentation or generated sentences. The core component of a Chatbot is Natural Language Understanding (NLU) which has been impressively improved by deep learning methods. But we often lack such properly built NLU modules and requires more time to build it from scratch for high quality conversations. This may encourage fresh learners to build a Chatbot from scratch with simple architecture and using small dataset, although it may have reduced functionality, rather than building high quality data driven methods. This research focuses on Named Entity Recognition (NER) and Intent Classification models which can be integrated into NLU service of a Chatbot. Named entities will be inserted manually in the knowledge base and automatically detected in a given sentence. The NER model in the proposed architecture is based on artificial neural network which is trained on manually created entities and evaluated using CoNLL-2003 dataset.*

"Keywords: Chatbot; Natural Language Understanding; Intent Classification; Named Entity Recognition"

## 1. Introduction

User machine conversation is a new engineering by using Natural Language Processing (NLP) approach and deep learning. One such conversation is the Chatbot, a virtual agent which interacts with user by providing meaningful answer. I have adopted this technology because of its increasing demand for business. Usage of Chatbots has been extensively amplified with the development of more and more Chatbots of different architecture. One of the earliest Chatbots is the ALICE bot which uses artificial intelligence mark-up language (AIML) [1], applying the technique of pattern matching based on user replies. After few decades, NLP techniques were evolved in order to improve the accuracy of existing models [2] [3]. Later it was found that deep learning models can give auspicious results while developing human machine interactions [4] [5] [6] [7]. Such interaction with machines has been significantly improved in the past few years with the combination of NLP and deep learning. Chatbot can be implemented in a number of ways using deep learning algorithms. A well-known deep learning method is Seq2Seq [8] [9] which can easily be adapted into a Chatbot. It is based on Recurrent Neural Network (RNN) [10] which have their internal memory to process sequence of inputs. Some Seq2Seq models use an advanced RNN called Long Short-Term Memory (LSTM) [11] in order to solve the problem of vanishing gradient.

Nowadays there are different Chatbots which can broadly be divided into two main categories and one of them is the task-oriented Chatbot, designed to achieve specific goals and can be used for definite business domains like online shopping, website guidance, restaurant reservations and booking airline tickets [12] [13] [6]. Task-oriented dialogue systems require a number of components where NLU is the main component to understand user's intent and recognize named entities. Such virtual agents function only for specific domain where user may ask question in natural language against available knowledge base and get response. That knowledge base is usually handcrafted in the files and can take a lot of time and money to create it from scratch. A large number of task-oriented dialogues are available in industry (Microsoft [1], Google [2], Amazon [3], IBM [4] and Facebook[5]) which provide web interfaces to create knowledge base and their focus is always on best implementation of NLU service. In recent years, many techniques have been employed for NLU services, like intent classification has achieved high results using Convolutional Neural Networks (CNN) [14] and combination of CNN with RNN [15]. NER is another challenging technique used in NLU service which researchers have worked on using various supervised, unsupervised, semi-supervised and active learning algorithms. The proposed NER model is a fully supervised algorithm like its previous work include support vector machines [16], maximum entropy models [17], decision trees [18], conditional random fields [19] and artificial neural networks [20]. This non-sequential model can outperform few existing NER systems and it can be very beneficial where we have to extract entities without taking into account the position of named entity in a given sentence. Researchers have also achieved higher accuracy of NER models by combining CNNs, LSTMs [21] and Conditional Random Fields (CRF) [22].

This paper presents simple architecture of a task-oriented Chatbot with a simple NER model which doesn't depend on position of named entities inside sequence. This proposed model is deployed in NLU service for the purpose of conversational service for a specific business domain, it can act as human beings while providing meaningful context but it is limited to improvement in its knowledge base at runtime. In current implementation, the knowledge base remains limited and Chatbot acts more likely a search engine where human searches a certain parameter and Chatbot will respond according to search parameter. Chatbots can also remember the previous conversation topic but still are unable to keep the whole conversation flow in memory and is not a scope of current implementation. The format of knowledge base used here is somewhat similar to RASA [6] format as described in section 2.1, it is manually created as done in different Chatbots rather than using web interfaces to create it. In the following sections, it is described how NLP is applied to the manually created knowledge base, comprising of two main characteristics i.e. intent classification and entity extraction. Although there are many NLU services that are used by researchers but I have developed the simple NLU service for understanding user's intent and extracting named entities in order to motivate fresh learners. Both are implemented using artificial neural networks, and this simple NLU function within the architecture of Chatbot may not be outperforming the existing NLU services.

---

## 2. Architectural Design

The core component of current Chatbot design is Natural language processing (NLP) which consists of two sub domains viz. natural language understanding (NLU) and natural language generation (NLG). NLU component is being used by a number of researchers and developers when creating conversational interfaces. In this research work, NLU component comprises of two components: The Intent Classifier and NER, developed using artificial neural networks and optimized by regularization and early stopping communications. Chatbots like ALICE [1] which were developed in the 20[th] century applied rule base approaches using artificial intelligence mark-up language (AIML). In recent years, many online dialogues are developed for the automated creation of knowledge bases.

In this paper work, knowledge base has been created manually in order to train artificial neural network models and is somewhat similar to RASA training data format. Knowledge base is created in JSON format which is divided into two main areas i.e. Input and Response. Input comprises of a list of messages that bot expects from the user. The other techniques. Then NLG module is the simplest case here which utilizes the classified intent and extracted entities, connects with knowledge base and responds to user with a meaningful answer. Above figure demonstrates the role of NLU and NLG within the proposed architecture of Chatbot.

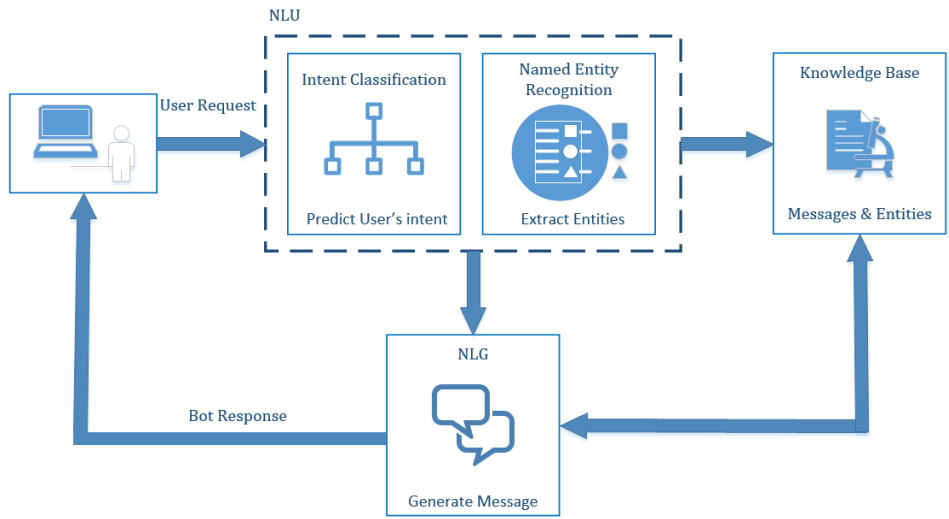

Figure 1: Proposed Architecture of Chatbot

### 2.1. Knowledge Base Creation

The intelligence of Chatbots is measured by the knowledge they have admittance to. Collection of such knowledge information data to train generative bot models is difficult to achieve human-like two keys along with the message are Intent and Entities. The intent describes what the messages are, for example, for a transportation network bot, the question: "I want to know taxi rate in Islamabad." has a "request_rate_taxi" intent. Entities are created only if these are included in the message otherwise it will be left as an empty array. Entities are important for the bot to understand what precisely a user is asking about.

```
{"Input":[{"Message": "How are you?","Intent":
"utter_greetings","Entities": []},
{"Message": "What is your name?","Intent":
"request_name","Entities": []}]}
{"Message": "What is taxi rate in Islamabad?":
"request_name",
"Entities": ["MISC":"taxi","LOC":"Islamabad"]}]}
```

"Response" section in knowledge base file contains list of messages that bot should respond according to user's intent. There can be multiple responses to the same question and randomly selected by the NLG module, because user may repeat the question. Information retrieval against NER makes use of next section in case user wants to extract information specific to entity. Like if user asks "I want to know the taxi rate in Islamabad", then "Islamabad" and "taxi" will be recognized as named entities and corresponding value will be extracted from the section "NE_Values", and NLG module will be capable of including this value in the response.

```
{"Response":[{
"utter_greetings": ["I am fine! What about you?","I
am fine. Thanks. ","Hello!I am good. How are you? "]
"request_name": ["I am Bot. What can I help
you?","I'm Bot. Do you need any help?"]},
{"NE_Values":[{"Islamabad":{ "taxi":{"Starting":"20
Rs./km","Minimum":"15
Rs./km"},"bike":{"Starting":"5 Rs./km","Minimum":"4
Rs./km"},
"business":{"Starting":"50Rs./km","Minimum":"40Rs.
/km"}}]}
```

## 2.2. Natural Language Understanding

There are several tools available online to build Chatbots without complex coding. These tools make use of NLU services in order to understand user's intent. Such online tools don't give any specific information about the used machine learning algorithms and the datasets on which these NLU services are trained initially [23]. One of such toolkit which is available as freeware is RASA NLU, this service is being used by a number of developers for building conversational interfaces. This paper describes the implementation of NLU service for intent classification and NER. Let's go through these in details:

### 2.2.1. Intent Classification:

After creation of knowledge base, next is to build the model which can understand user's intent so that when user asks any question like "What is your name?" then bot can understand it as "request_name" intent. Our training data is available in JSON format somewhat similar to RASA, having list of chat messages as unstructured text. Before building a model we need to pre-process the data and convert it to a structured format which is readable by machine learning model. It will be done as described below.

#### 2.2.1.1. Preprocessing:

In the first step, sentence is broken down into individual words called tokens. Then converted these tokens to lower case and removed the punctuations, numbers and white spaces. This step doesn't take union of these tokens because if a word is repeated for given sentence then it will be counted as twice when creating bag of words model. These tokens can be analyzed by a number of techniques like punctuations, abbreviations and periods can also be dealt with. Stop words removal is not a component of NLP operations here and it will be done when creating a model for NER. Then lemmatization, removal of inflectional endings, will ensure that words having the same meaning will be brought into the same form e.g. the word "say" and "says" have the same meaning and should not be repeated when creating vectors for different sentences. In the last step of pre-processing, the derived words are reduced to their base form a process called stemming. Like the word "played" will be reduced to "play".

#### 2.2.1.2. Bag of Words:

The target is the conversion of the text into numbers because machine learning algorithms can only understand numerical data. For that purpose, a list is created comprising of all the unique words which are found in the list of messages. Then count the words for each message ("document") and a matrix is constructed called term document matrix as shown below. Term document matrix can also be constructed by multiplying the term frequency with inverse document frequency. This measure will tell us the importance of a word to a document and is not a scope of this paper.

Here each row represents the message with number of words contained in it. This can also be called as bag of words model. We are in the last stage of pre-processing now where user defined intent is inserted at the end of term document matrix after encoding it into unique numbers.

Table 1: Term Document Matrix

| Index | please | are | need | which | how | taxi | . | . | . | rate | you | your | in | name | what | intent |
|---|---|---|---|---|---|---|---|---|---|---|---|---|---|---|---|---|
| 1 | 0 | 1 | 0 | 0 | 1 | 0 | . | . | . | 0 | 1 | 0 | 0 | 0 | 0 | utter_greetings |
| 2 | 0 | 0 | 0 | 0 | 0 | 0 | . | . | . | 0 | 0 | 1 | 0 | 1 | 1 | request_name |
| 3 | 0 | 0 | 0 | 0 | 0 | 1 | . | . | . | 1 | 0 | 0 | 1 | 0 | 1 | request_rate |
| 4 | 1 | 1 | 0 | 1 | 0 | 0 | . | . | . | 0 | 0 | 0 | 0 | 0 | 0 | request_docs |

*2.2.1.3. Artificial Neural Network Model:*

After pre-processing, our data is ready to be fed into the classifier model. For intent classification, we are using multi-class artificial neural network with 2 hidden layers, determined after hyper-parameter tuning. Number of neurons in input layer is equal to number of terms in term document matrix while Softmax layer neurons is equal to number of intents available in knowledge base. Softmax layer will produce likelihoods for each intent, which sum up to one.

Next train/test split is performed on pre-processed data in such a way that distribution of intents remains same in both datasets as multiple messages are created in knowledge base for each of the intent. Model is trained with training data for certain number of epochs, fitting the set of features against user defined intent. Then we will have statistical model which will be able to take the user message and predict the user's intent. The accuracy of the model would improve with larger datasets. First model is tested with training data and then over testing data in order to avoid over/under fitting problems.

User's message will go through the same pre-processing phase as done for training the list of messages. For example, if user says, *"How are you?"* then pre-processing would result in the sentence vector shown in Table 2.

This vector will be passed as input to the artificial neural network model for predicting the user's intent, Softmax layer at output would return probability for each of the intents the model has been trained on. We look for the maximum probability to be above a certain threshold which will be decoded in order to arrive at the actual intent. If the maximum probability is below threshold then bot would not be clever enough to understand consumer's intent.

*2.2.2. Named Entity Recognition*

Entity extraction is very important for NLU services. Multiple statistical models are available to extract entities like NER Spacy and NER Duckling, commonly used to extract pre-trained entities like name, location, date etc. while NER CRF can be used for custom entities. In this work entities stored in training JSON file as key value pairs, are split into characters and generate a vector representation for all the available entities while using characters count as shown in Table 3.

This dataset is divided into three parts i.e. train, test and validation. The possible entity types which are available in three datasets are name, locations, organizations and miscellaneous. Class of each entity is inserted as the last column of the characters count matrix, this predictive column needs to be encoded into numbers so that it can be read by our ANN model. Unique characters found in the named entities will be used as features for this model. Once the data is converted to such format that it is supported by our ANN model, then model is trained with train set for a number of iterations while validation set is used when to stop training. There are two hidden layers and output layer having 4 unit which return sequence of vectors containing Softmax probability for every input vector.

Table 2: Sentence Vector

| please | are | need | which | how | taxi | . | . | . | rate | you | your | in | name | what | intent |
|---|---|---|---|---|---|---|---|---|---|---|---|---|---|---|---|
| 0 | 1 | 0 | 0 | 1 | 0 | . | . | . | 0 | 1 | 0 | 0 | 0 | 0 | ? |

Table 3: Characters Count Matrix

| Index | Entity | a | b | c | d | e | f | g | . | . | . | u | v | w | x | y | z | Entity type |
|---|---|---|---|---|---|---|---|---|---|---|---|---|---|---|---|---|---|---|
| 1 | Islamabad | 3 | 1 | 0 | 1 | 0 | 0 | 0 | . | . | . | 0 | 0 | 0 | 0 | 0 | 0 | LOC |
| 2 | Karachi | 2 | 0 | 1 | 0 | 0 | 0 | 0 | . | . | . | 0 | 0 | 0 | 0 | 0 | 0 | LOC |
| 3 | Taxi | 1 | 0 | 0 | 0 | 0 | 0 | 0 | . | . | . | 0 | 0 | 0 | 1 | 0 | 0 | MISC |
| 4 | HiveWorx | 0 | 0 | 0 | 0 | 1 | 0 | 0 | . | . | . | 0 | 1 | 1 | 1 | 0 | 0 | ORG |

Table 4: Characters Count Matrix for given sentence

| word | a | b | c | d | e | f | g | . | . | . | u | v | w | x | y | z | entity |
|---|---|---|---|---|---|---|---|---|---|---|---|---|---|---|---|---|---|
| taxi | 1 | 0 | 0 | 0 | 0 | 0 | 0 | . | . | . | 0 | 0 | 0 | 1 | 0 | 0 | ? |
| rate | 1 | 0 | 0 | 0 | 1 | 0 | 0 | . | . | . | 0 | 0 | 0 | 0 | 0 | 0 | ? |
| islamabad | 3 | 1 | 0 | 1 | 0 | 0 | 0 | . | . | . | 0 | 0 | 0 | 0 | 0 | 0 | ? |

In order to predict the entities, the user's utterance is tokenized and stop words are eliminated because stop words are not in entities, then the remaining words are split into characters as entity recognition is performed on each word in the document like if user says "What is the taxi rate in Islamabad?" then generated matrix would be as shown above in Table 4. The resulting matrix is passed into the entity extraction model which supports rows of vectors as input. Since our classifier is trained on four entity types, it would return Softmax probabilities for each type. We look for the maximum probability to be above a certain threshold for each vector and model is capable of predicting one or more entities for a given sentence.

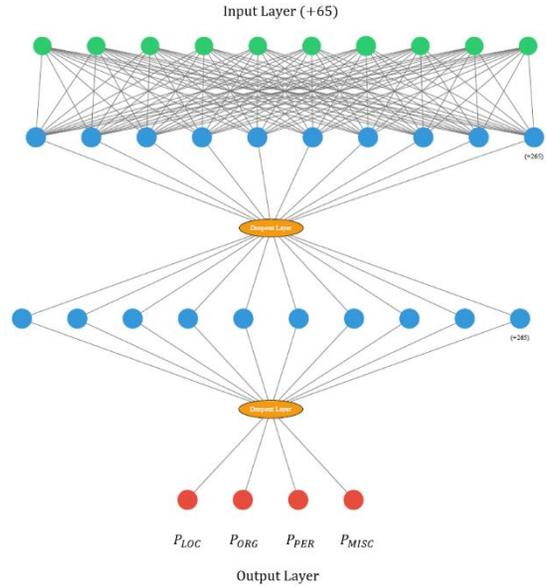

Figure 2: ANN model to extract entities

## 3. Experiments and Results

This paper describes simple architecture of a Chatbot employed with intent classification and named entity recognition models which can be trained on knowledge base created for specific business domain and can be adapted to any business domain by creating the knowledge base accordingly. Artificial Neural Network is potentially general architecture for the applications that include classification problem. The proposed architecture of Chatbot comprises of two classification problems (intent and entity) which are addressed by two artificial neural network models and perform well on question answer data that is restricted to a specific domain, but unable to perform well on large information based questions. This research is focused on implementation and evaluation of Named Entity Recognition model within the architecture of Chatbot and it is very important for other applications like Machine Translation, Question Answering and Information Retrieval. Model evaluation shows that it can outperform one of the existing system and it can be beneficial than sequence models in situations where input is not a sequence of words and can be in the form of list of words in any order. In order to evaluate the model, I have used CoNLL-2003 English dataset which is freely available which was expressed in the shared task of the Conferences on Computational Natural Language Learning [24]. This is a usually considered dataset with 4 typical types of entity: names, organizations, locations and miscellaneous names. The results of the proposed model are comparable with one of the existing systems while outperforming another one.

The proposed model is trained with CoNLL-2003 English train set while accuracy with the validation set is used to stop training when accuracy is maximum. Then test set is used to evaluate the trained NER model. Classification report and confusion matrix show that results are comparable with Memory-Based Named Entity Recognition using Unannotated Data [25] and better than NER model which was implemented using Long Short-Term memory [26]. Although this system doesn't outperform latest NER systems but its benefit is that it doesn't depend on sequence of words (sentence) as input and works well in situations where dataset is specific to a business domain.

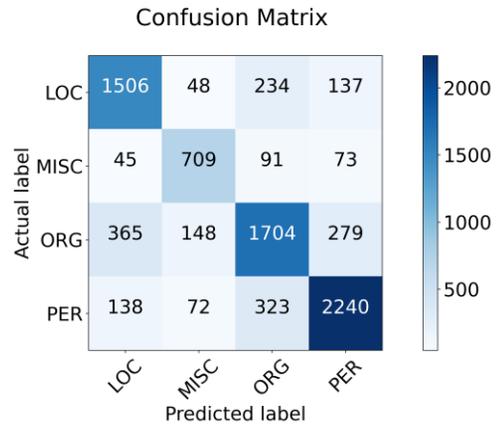

Figure 3: Confusion matrix on test set

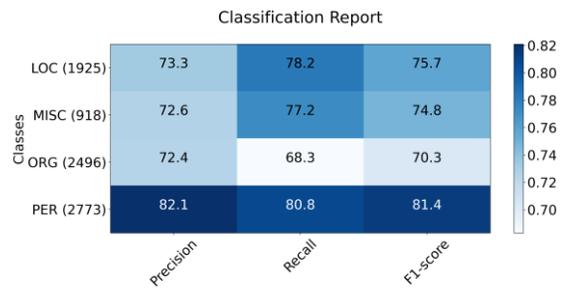

Figure 4: Classification report on test set

Table 5: Results Comparison with English test set

| English test set | Precision | Recall | F1-Score |
|---|---|---|---|
| Proposed NER Model | 75.96% | 75.92% | 75.89% |
| [25] | 75.84% | 78.13% | 76.97% |
| [26] | 69.09% | 53.26% | 60.15% |
| Baseline | 71.91% | 50.90% | 59.61% |

This NER model can be adapted by any business domain like it is used for Transportation Network Company, the model is trained with maximum of 780 named entities and doesn't require high computational cost and memory. One of the disadvantage of proposed NER model is that it can treat some named entities as same vector like two persons "Amna" and "Anam" will be treated as same person being converted to same vectors, ignoring the order of characters for each word. Although this NER model can work with a maximum accuracy for such Chatbot but a lot of work still can be merged for user's intents like it may not answer very common real life questions and can be included in knowledge base. It can keep previous user's intent in memory but can't keep the whole conversation flow in memory, like resulting chat below shows that questions are independent of each other. Another limitation is that it would treat two sentences as the same user's intent like "He is traveling to London." and "Is he traveling to London?" will appear with same vector.

be annotated with different tags when creating knowledge base.

Performance of NER model can significantly be improved when trained on a larger dataset. Model evaluation with CoNLL-2003 validation set would give results better than test set.

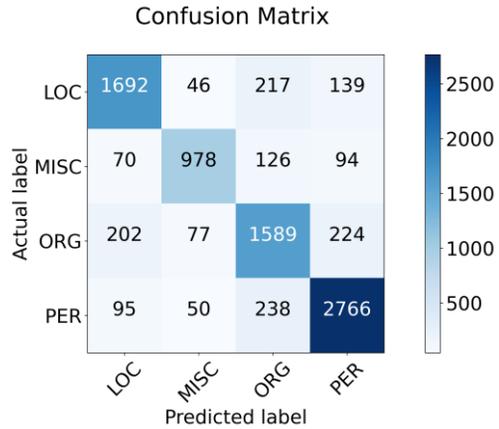

Figure 5: Confusion matrix on validation set

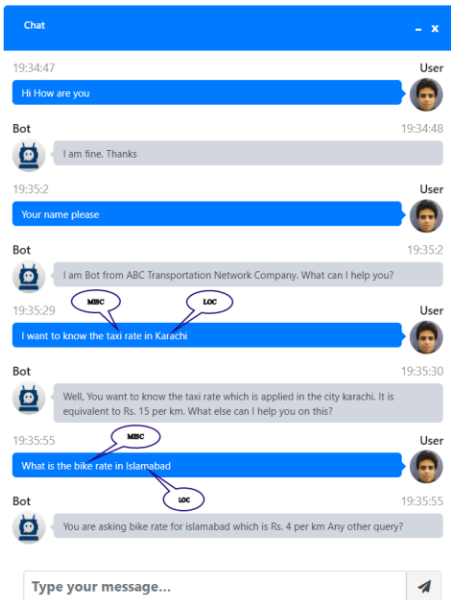

Figure 4: Sample Chat with predicted entities

For intent classification model, we ignore order of words and also remove punctuations when generating bag of words model. Intent classification model is evaluated with train/test split and accuracy of 89% is achieved on test set which can further be improved by using larger datasets. The questions and named entities which have similar vector representations can

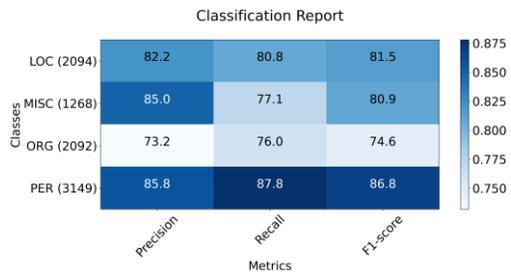

Figure 6: Classification report on validation set

Table 6: Evaluation metrics on validation set

| English validation set | Precision | Recall | F1-Score |
|---|---|---|---|
| Proposed NER Model | 81.74% | 81.65% | 81.66% |

## 4. Conclusions and Future Work

In this paper, a simple Chatbot design has been presented with proposed NER model which can accelerate interest of fresh learners in the field of artificial intelligence (AI) or if someone is looking to venture into this field then it could be a good starting point. This architecture is useful for simple tasks and which do not require much of memory and processing. It uses training data in the form of JSON file having attribute-value pairs which are self-describing and easily editable. The data format is always very important for any artificial learning application. This conversational agent is based on artificial neural networks which can outperform the performance of other machine learning algorithms when using datasets of larger sizes. The proposed system is used for a transportation network company and is adaptable enough to be used for others businesses like online shopping, web page guidance, hotel reservations, booking airline tickets etc., providing service to its customers. The current implementation has a positive influence on fresh learners' motivation and desire to understand neural networks and their usage in AI applications.

The proposed system can be used in various conversation contexts and can be easily modified where artificial neural networks can be replaced by memory networks in order to improve accuracy and make it more robust and efficient. More conversation content can be added in order to improve the accuracy of models, and more keys, like synonyms, can be included which make the system more efficient and improve this text-based Chatbot by providing more communicative features.